\documentclass[conference]{IEEEtran}
\IEEEoverridecommandlockouts
\usepackage{cite}
\usepackage{amsmath,amssymb,amsfonts}
\usepackage{algorithmic}
\usepackage{graphicx}
\usepackage{textcomp}
\usepackage{xcolor}
\usepackage{url}            
\usepackage{booktabs}       
\usepackage{amsfonts}       
\usepackage{nicefrac}       
\usepackage{microtype}      
\usepackage{color}
\usepackage{graphicx}
\usepackage{float}
\usepackage{multirow}
\usepackage{adjustbox}
\usepackage[normalem]{ulem}
\usepackage{algorithmic}
\usepackage{algorithm}
\usepackage{bbm}
\usepackage{wrapfig}
\useunder{\uline}{\ul}{}
\usepackage{balance}
\usepackage{stfloats}
\usepackage[bookmarks=false]{hyperref}

\hypersetup{
    colorlinks=true,
    linkcolor=black,
    filecolor=magenta,      
    urlcolor= black,
    citecolor = black,
}
\hypersetup{draft}
\usepackage[inline]{enumitem}

\graphicspath{ {figures/} }

\newcommand{\bm}[1]{\boldsymbol{#1}}

\def\BibTeX{{\rm B\kern-.05em{\sc i\kern-.025em b}\kern-.08em
    T\kern-.1667em\lower.7ex\hbox{E}\kern-.125emX}}

\begin{document}

\makeatletter
\newcommand{\linebreakand}{%
  \end{@IEEEauthorhalign}
  \hfill\mbox{}\par
  \mbox{}\hfill\begin{@IEEEauthorhalign}
}
\makeatother

\title{Extendable and invertible manifold learning with geometry regularized autoencoders
\thanks{$^*$Equal contributions; $^\dagger$ Equal contributions, corresponding authors. This research was partially funded by an IVADO (l'Institut de valorisation des donn\'{e}es) 
Undergraduate introduction to research scholarship [\emph{S.M.}]; IVADO Professor research funds, and NIH grant R01GM135929 [\emph{G.W}]. The content provided here is solely the responsibility of the authors and does not necessarily represent the official views of the funding agencies.}
}

\author{\IEEEauthorblockN{Andr\'{e}s F. Duque$^*$}
\IEEEauthorblockA{\textit{Dept. of Math. \& Stat.} \\
\textit{Utah State University}\\
Logan, UT, USA \\
andres.dc@aggiemail.usu.edu}
\and
\IEEEauthorblockN{Sacha Morin$^*$}
\IEEEauthorblockA{\textit{Dept. of Comp. Sci.} \\
\textit{Univ. de Montr\'{e}al; Mila}\\
Montreal, QC, Canada \\
sacha.morin@umontreal.ca}
\and 
\IEEEauthorblockN{Guy Wolf$^\dagger$}
\IEEEauthorblockA{\textit{Dept. of Math. \& Stat.} \\
\textit{Univ. de Montr\'{e}al; Mila}\\
Montreal, QC, Canada \\
guy.wolf@umontreal.ca}
\and
\IEEEauthorblockN{Kevin Moon$^\dagger$}
\IEEEauthorblockA{\textit{Dept. of Math. \& Stat.} \\
\textit{Utah State University}\\
Logan, UT, USA \\
kevin.moon@usu.edu}
}

\IEEEoverridecommandlockouts
\IEEEpubid{\makebox[\columnwidth]{978-1-7281-6251-5/20/\$31.00~\copyright2020 IEEE \hfill} \hspace{\columnsep}\makebox[\columnwidth]{ }}

\maketitle

\IEEEpubidadjcol

\begin{abstract}
A fundamental task in data exploration is to extract simplified low dimensional representations that capture intrinsic geometry in data, especially for  faithfully visualizing data in two or three dimensions. Common approaches to this task use kernel methods for manifold learning. However, these methods typically only provide an embedding of fixed input data and cannot extend to new data points.  Autoencoders have also recently become  popular for representation learning. But while they naturally compute feature extractors that are both extendable to new data and invertible (i.e., reconstructing original features from latent representation), they have limited capabilities to follow global intrinsic geometry compared to kernel-based manifold learning. We present a new method for integrating both approaches by incorporating a geometric regularization term in the bottleneck of the autoencoder. Our regularization, based on the diffusion potential distances from the recently-proposed PHATE visualization method, encourages the learned latent representation to follow intrinsic data geometry, similar to manifold learning algorithms, while still enabling faithful extension to new data and reconstruction of data in the original feature space from latent coordinates. We compare our approach with leading kernel methods and autoencoder models for manifold learning to provide qualitative and quantitative evidence of our advantages in preserving intrinsic structure, out of sample extension, and reconstruction. Our method is easily implemented for big-data applications, whereas other methods are limited in this regard. 
\end{abstract}


\section{Introduction}
\label{sec:introduction}
The high dimensionality of modern data introduces significant challenges in descriptive and exploratory data analysis. These challenges gave rise to extensive work on dimensionality reduction aiming to provide low dimensional representations that preserve or uncover intrinsic patterns and structures in processed data. A common assumption in such work is that high dimensional measurements are a result of (often nonlinear) functions applied to a small set of latent variables that control the observed phenomena of interest, and thus one can expect an appropriate embedding in low dimensions to  recover a faithful latent data representation. While classic approaches, such as principal component analysis (PCA)~\cite{pearson1901} and classical multidimensional scaling (MDS)~\cite{cox2008multidimensional}, construct linear embeddings, more recent attempts mostly focus on nonlinear dimensionality reduction. These approaches include manifold learning kernel methods and deep learning autoencoder methods, each with their own benefits and deficiencies.

Kernel methods for manifold learning cover some of the most popular nonlinear dimensionality reduction methods, dating back to the introduction of Isomap~\cite{tenenbaum2000global} and Locally Linear Embedding (LLE)~\cite{roweis2000nonlinear}. These two methods proposed the notion of data manifolds as a model for intrinsic low dimensional geometry in high dimensional data. The manifold construction in both cases is approximated by a local neighborhood graph, which is then leveraged to form a low-dimensional representation that preserves either pairwise geodesic distances (in the case of Isomap) or local linearity of neighborhoods (in LLE). The construction of neighborhood graphs to approximate manifold structures was further advanced by Laplacian eigenmaps~\cite{belkin2002laplacian} and diffusion maps \cite{coifman2006diffusion}, together with a theoretical framework for relating the captured geometry to Riemannian manifolds via the Laplace-Beltrami operators and heat kernels. These approaches that, until recently, dominated manifold learning can collectively be considered as spectral methods, since the embedding provided by them is based on the spectral decomposition of a suitable kernel matrix that encodes (potentially multiscale) neighborhood structure from the data. They are also known as kernel PCA methods, as they conceptually extend the spectral decomposition of covariance matrices used in PCA, or that of a Gram (inner product) matrix used in classic MDS.

Recent work in dimensionality reduction has focused on visualization for data exploration. Spectral methods are generally unsuitable for such tasks because, while their learned representation has lower dimension than the original data, they tend to embed data geometry in more dimensions than can be conveniently visualized (i.e., $\gg 2$ or 3). This is typically due to the orthogonality constraint and linearity of spectral decompositions with respect to the initial dimensionality expansion of kernel constructions~\cite{moon2019visualizing,haghverdi2016diffusion}. This led to the invention of methods like t-SNE (t-Distributed Stochastic Neighbor Embedding) \cite{maaten2008visualizing}, UMAP (Uniform Manifold Approximation and Projection) \cite{UMAP2018}, and PHATE (Potential of Heat-diffusion for Affinity-based Transition Embedding) \cite{moon2019visualizing}. These methods embed the data by preserving pairwise relationships between points and can  be viewed as generalizations of metric and non-metric MDS. These methods and their extensions have been used in applications such as  single cell genomics~\cite{moon2019visualizing,amir2013visne,sgier2016flow,macosko2015highly,becht2019dimensionality,zhao2020single,karthaus2020regenerative,baccin2019combined}, visualizing time series~\cite{duque2019}, visualizing music for a recommendation system~\cite{van2013deep}, and analyzing the internal representations in neural networks~\cite{2019Gigante,horoi2020low}. However, these and spectral methods typically provide fixed latent coordinates for the input data. Thus, they do not provide a natural embedding function to perform out-of-sample extension (OOSE). This shortcoming is usually tackled by employing geometric harmonics \cite{coifman2006geometric}, Nystr\"om extension \cite{williams2001using}, or a landmark approach \cite{vladymyrov2013locally}.

In contrast, Autoencoders (AEs) are a different paradigm for non-linear dimensionality reduction. First introduced in \cite{rumelhart1986learning}, this non-convex and parametric approach has gained more attention in recent years, especially due to the computational and mathematical advances in the field enabling the efficient implementation and training of neural networks. In contrast to kernel methods, AEs learn a parametric function and are thus equipped with a natural way to perform OOSE, as well as an inverse mapping from the latent  to the input space. Despite these  properties, AEs usually fail to accurately recover the geometry present in the data. This not only restricts their desirability to perform exploratory data analysis, e.g. via low-dimensional visualization, but  can also lead   to bad reconstructions over certain regions of the data, as we show in this work.

Motivated by the complementary advantages provided by AE and kernel methods, we introduce Geometry regularized autoencoders (GRAE), a general framework which splices the well-established machinery from kernel methods to recover a sensible geometry with the parametric structure of AEs. Thus we gain the benefits of both methods, furnishing kernel methods with efficient OOSE and inverse mapping, and providing the autoencoder with a geometrically driven representation. To achieve this, GRAE introduces a regularization on the latent representation of the autoencoder, guiding it towards a representation previously learned by a kernel method. 
In this paper we focus our presentation using PHATE \cite{moon2019visualizing} as our preferred method for finding a sensible geometry. 
Nevertheless, our general formulation can also be easily implemented with other methods such as UMAP. 

Our main contributions are as follows:
\begin{enumerate*}
    \item We present a general approach to leverage traditional dimensionality reduction and manifold learning methods, providing them with a natural OOSE and an invertible mapping.   
    \item By introducing the geometric regularization, the reconstruction error of the Autoencoder decreases in many cases, suggesting that  the previously learned geometry leads to a better representation for reconstruction. 
    \item We show how our method can be implemented in a big data setting, which is a common limitation  in  related work.
\end{enumerate*}

\section{Related Work}
\label{sec:related_work}

\paragraph{Manifold learning}

Manifold learning methods for dimensionality reduction typically assume data lie on a low dimensional manifold $\mathcal{M}$ immersed in the high dimensional ambient space. Therefore they aim to map points from $\mathcal{M}$ to a low dimensional Euclidean space that encodes or reveals its intrinsic geometry. However, in practice, such methods only consider a finite set of data points $x_{1}, \ldots, x_{n} \in \mathbbm{R}^\mathcal{D}$ (for $\mathcal{D}$ dimensional ambient space), assumed to be sampled from $\mathcal{M}$, and optimize a fixed set of low dimensional points $y_{1}, \ldots, y_{n} \in \mathbbm{R}^d$ (for $d \ll D$) such that the Euclidean relations between pairs $(y_i,y_j)$ will reflect intrinsic nonlinear  relations between the corresponding $(x_i,x_j)$. Recent manifold learning kernel methods typically follow the framework introduced in \cite{snepaper} and further extended by t-SNE \cite{maaten2008visualizing}, which are themselves generalization of the metric MDS algorithm, whereby the coordinates in the latent space are optimized by gradient descent  to recreate the pairwise similarities (as defined by a kernel) in the input space. Intuitively, the use of a kernel which outputs high similarities for close neighbors enables the capture of the curvature of the underlying manifold in the ambient space. t-SNE, for instance, uses normalized Gaussian similarities in the input space and t-distributed similarities in the latent space. The embedding is optimized so as to minimize the Kullback-Leibler divergence between both distributions.

UMAP \cite{UMAP2018} was introduced as an improvement of t-SNE, with claims of improved preservation of global features and better run times. Specifically, the cost function of t-SNE is replaced by cross-entropy and similarities between objects in the input space are computed based on the smooth nearest neighbor distances, that is: $v_{j|i}= exp((-d(x_i, x_j) - p_i) /\sigma_i)$, where $p_i$ is the distance between $x_i$ and its nearest neighbor, $\sigma_i$ is the bandwidth and $d$ is a distance, not necessarily Euclidean. In contrast with t-SNE, UMAP does not normalize similarities and relies on an approximation of the neighborhoods using the Nearest-Neighbor-Descent algorithm of \cite{dong2011efficient}. The UMAP implementation further distinguishes itself from t-SNE by not restricting the embedded space to two or three dimensions.

Recently, the claim that UMAP improves over t-SNE in preserving global structure has been challenged in \cite{kobak2019umap}. The authors attribute the better global structure commonly obtained by UMAP to the  differences between both methods in the initialization procedure. Typically t-SNE  uses a random initialization, whereas UMAP uses Laplacian eigenmaps as its starting point. They showed that nearly identical results can be obtained by also initializing t-SNE with Laplacian eigenmaps.

\paragraph{Autoencoders}
 PCA naturally provides an extendable and (approximately) invertible embedding function, but falls short in capturing  non-linear mappings. To see this, recall that the embedding function is constructed using a matrix $M$ with orthogonal columns consisting of principal components (PCs) such that $M M^T$ is a projection operator, thus acting as an identity on a hyperplane spanned by the PCs. Then, the  embedding is given by the linear function $\hat{x} =  M^T x$ and its inverse (on the embedded space) is given by $M \hat{x} = M M^T x \approx x$, where the approximation quality (i.e. reconstruction error) serves as the optimization target for computing the PCs in $M$. 
 
 To generalize this construction to  nonlinear embedding functions over a data manifold $\mathcal{M}$, autoencoders (AEs) replace $M$ by an encoder function $f: \mathcal{M} \xrightarrow{} \mathbbm{R}^d$ and $M^T$ by a decoder function $f^{\dagger}: \mathbbm{R}^d \xrightarrow{} \mathcal{M}$, which is an approximate inverse of $f$. Both functions are  parametrized by a neural network and trained via a reconstruction loss to ensure the composite function $f^{\dagger} \circ f$ acts as an identity on data sampled from $\mathcal{M}$. By considering datasets in matrix notation (i.e., with rows as datapoints), the AE optimization is generally formulated as
\begin{align}
\label{eq:AE}
\operatorname*{arg\ min}_{f,f^{\dagger}}\mathcal{L}(f,f^{\dagger}) = \mathcal{L}_{r}(X, f^{\dagger}(f(X))),
\end{align}
where $f,f^{\dagger}$ are understood to be applied separately to each row in their input matrix (yielding corresponding output data points organized in matrix form), and $\mathcal{L}_{r}$ denotes a loss function that measures the discrepancy between the original and reconstructed data points (commonly MSE). It is common to select $d < \mathcal{D}$, forcing the autoencoder to find a representation in latent codes of dimension $d$  while retaining as much information for reconstruction as possible. In this case the autoencoder is  \textit{undercomplete}. Under this formulation, instead of learning new coordinates for the input data, we learn an embedding function $f$ and an inverse function $f^{\dagger}$. If $f$ is a linear function, the network will project onto the same subspace spanned by the principal components in PCA~\cite{baldi1989neural}.

\paragraph{Regularized autoencoders}
The vanilla AE formulation in (\ref{eq:AE}) has been extended for many purposes by adding regularization as a prior on the function space of $f$ and $f^{\dagger}$. Denoising AEs \cite{vincent2008extracting} are widely used to find good latent representations and perform feature extraction, exploiting the flexibility provided by neural networks (e.g. \cite{supratak2014feature,liu2014feature}). Contractive autoencoders (CAE) \cite{rifai2011contractive} penalize the Frobenius norm of the Jacobian of the encoder $f$, encouraging a more robust representation around small perturbations from the training data. When dealing with a high dimensional latent space (e.g., the \textit{overcomplete} case), sparse autoencoders \cite{ng2011sparse} are particularly useful, introducing a sparsity constraint that forces the network to learn significant features in the data. Extensions to produce generative models, such as variational autoencoders (VAE) \cite{kingma2013auto}, regularize the latent representation to match a tractable probability distribution (e.g., isotropic multivariate Gaussian), from which it is possible to sample over a continuous domain to generate new points.
 
More closely related to our work, some attempts to impose geometrically driven regularizations on the latent space have been proposed over the past two decades. Stacked Similarity-Aware Autoencoders, for instance, enforce a cluster prior based on pseudo-class centroids to increase subsequent classification performance \cite{chu2017stacked}.
 
 A relatively new implementation called Diffusion Nets~\cite{mishne2019diffusion} encourages the AE embedding to learn the geometry from Diffusion Maps (DM)~\cite{coifman2006diffusion}, a manifold learning algorithm. This approach combines an MSE loss in the embedding coordinates with the so-called eigenvector constraint to learn the diffusion geometry. Diffusion Nets inherits some of the inherent issues of Diffusion Maps. Perhaps most importantly, they inherit its inability as a spectral method to ensure significant representation of the data on a fixed lower dimension, due to the natural orthogonality imposed among the diffusion coordinates~\cite{moon2019visualizing}. Therefore,  effective use of Diffusion Nets may require the network architecture 
to be determined from the numerical rank of the diffusion affinity kernel used in DM. This, in turn, would limit the capabilities of this approach in data exploration (e.g., visualization), while in contrast PHATE (and UMAP) can specifically optimize a chosen dimension (e.g., 2D or 3D). Moreover, as a spectral  method, DM itself\footnote{We note that the DM runtime (relative to UMAP) is equivalent to that of  Laplacian eigenmaps  reported in~\cite{UMAP2018}, as the algorithmic difference between these spectral methods is negligible~\cite{coifman2006diffusion}.} tends to be more computationally expensive than PHATE (which we use in this work, see Sec. \ref{sec:GRAE}) or UMAP~\cite{moon2019visualizing,UMAP2018}. 


The formulation of Diffusion Nets is closely related to Laplacian autoencoders (LAE)~\cite{jia2015laplacian} and Embedding with Autoencoder Regularization (EAER)~\cite{yu2013embedding}. Both of these methods include a regularization term that penalizes inaccurate preservation of neighborhood relationships in the original space:
\begin{multline}
\operatorname*{arg\ min}_{f,f^\dagger}\mathcal{L}(f,f^\dagger) = \mathcal{L}_{r}(X, f^\dagger(f(X)))\\ + \lambda\sum_{i < j}^{n}\mathcal{L}'(f(x_{i}),f(x_{j}), \phi_{ij}).
\label{lae}
\end{multline}


For instance, in EAER the second term can be the classical MDS objective $\sum_{i < j}^{n} (\|f(x_{i})- f(x_{j})\| - d_{ij})^{2}$ where $d_{ij}$ is a given distance computed in the input space. It can also be margin-based, where embedding distances between neighbors are penalized if not 0, while distances between non-neighbors are penalized if not above a certain margin. Finally, $\mathcal{L'}$ can take the form  $\sum_{i < j}^{n} \|f(x_{i})- f(x_{j})\|^{2}\phi_{ij}$, where  $\phi_{ij}$ are computed using the weighted edges in an adjacency graph. This gives the objective function of Laplacian eigenmaps. LAE employs a similar loss term, but goes further by adding a second order term including the Hessian of $f$.


Another interesting approach derived from manifold learning methods can be found in \cite{pai2019dimal}. The authors aimed to replicate Isomap's objective function by using a SIAMESE network architecture trained over the pairwise geodesic distances. Their method, however, scales quadratically in the number of landmarks selected for training.


Recently, topological autoencoders (TAE)  \cite{moor2019topological} were proposed which include a regularization based  on a topological signature of the input data. The topology of a manifold can  be characterized by \textit{homology groups}, which, depending on their dimension, represent various topological features such as the number of disconnected components or the number of cycles on the manifold that cannot be deformed into each another. When data are sampled from a manifold, such topological features can be approximately derived from a $\varepsilon$-ball neighborhood graph of the data. Persistent homology \cite{edelsbrunner2008persistent,barannikov1994framed} was introduced as a means to identify the topological signature of manifolds based on how long topological features persist when progressively increasing the $\varepsilon$-ball of the neighborhood graph up to a point where the graph is fully-connected (topological features with a short $\varepsilon$ lifespan are attributed to noise). TAE thus penalizes discrepancies between the topological signatures of the input space and the latent space.

\paragraph{Out of sample extension}
 Manifold learning algorithms are typically based on the eigendecomposition of a kernel matrix (Diffusion Maps) or a stochastic optimization of the latent coordinates (metric MDS, UMAP). Therefore, in constrast to neural networks, they do not provide a general embedding function that operates on, or provides a representation of, the entire manifold $\mathcal{M}$. Thus these methods are not applicable to arbitrary input points in $\mathbbm{R}^\mathcal{D}$, which we would ideally want to project onto the learned manifold.  To address this shortcoming, a parametric version of t-SNE using a neural network approach was proposed in \cite{van2009learning} where a multi-step training procedure is used to optimize the t-SNE objective with a neural network. 
 

Another well-known solution to the lack of OOSE is the Nystr\"{o}m extension \cite{bengio2004out} and its improvements, such as geometric harmonics \cite{coifman2006geometric}, which  approximate an empirical function over new points using linear combinations of the eigenvectors of a kernel matrix computed on the training set. Let $X = \{x_1, ... , x_m\}$ be the training set used to compute the initial kernel matrix $K$ with kernel function $k(\cdot, \cdot)$. Then a new point $x'$ can be extended to the learned latent space using the eigenvectors $\psi_i$ of $K$ with eigenvalues $\lambda_i$ as follows: $\hat{\psi_i}(x') \approx \frac{1}{\lambda_i}\sum\limits_{j=1}^{m}k(x_j, x')\psi_j(x_j)$.

One can thus project a function on the eigenvectors of $K$ and then extend to new points using the new approximated eigenvectors. However, this approach has several drawbacks \cite{mishne2019diffusion}. Given $n$ training points (resulting in $K$ being $n\times n$), extending a function to $m$ points requires us to compute $m$ new kernel rows leading to a time complexity of $\mathcal{O}(nm)$. Furthermore,  the empirical function must be within the interpolation range of the kernel, which requires bandwidth tuning. 


Other methods,  perform OOSE by a linear combination of training points (the ``landmarks'') close to the new points in the input space  \cite{vladymyrov2013locally}, as in PHATE~\cite{moon2019visualizing} and landmark MDS~\cite{silva2003global}.  UMAP takes a similar approach to OOSE by initializing latent coordinates of new points in the latent space based on their affinities with training points in the input space. The new layout is then optimized by gradient descent following the same optimization scheme as the main algorithm, using the embeddings of training points as reference. 

All of these approaches require the training points, or a subset, to be stored in memory with their embeddings as a proxy for the target function, which can quickly become inconvenient given a large dataset or lead to a loss in embedding quality due to subsampling or the use of landmarks. Moreover, they do not provide a straightforward approximation of the inverse function, which is critical in assessing how well the information is preserved in the embedded space. As such, they are not directly comparable to GRAE and other AE based models, which present a native approximation of the inverse and only need to store the weights and biases of the network to perform OOSE, thus having memory requirements independent of the size of the training set.

\section{Geometry-regularized autoencoder}
\label{sec:GRAE}
\subsection{Learning embedding functions}

In this work, we aim to learn a data manifold geometry to find an appropriate embedding function $f : \mathcal{M} \to \mathbbm{R}^d$, rather than just a fixed point-cloud embedding. This contrast  can be seen, for example, by considering the classic PCA and MDS methods. While both of these methods can be shown analytically to extract equivalent (or even the same) linear embeddings from data, MDS only assigns coordinates to fixed input points (similar to the described manifold learning methods), while PCA provides an embedding function (albeit linear) defined by a projection operator on principal components. Here, we aim to establish a similar equivalence in nonlinear settings by providing an alternative to popular manifold learning approaches that constructs an embedding function (as a nonlinear successor of PCA) and also yields representations that capture an intrinsic geometry similar to that of established kernel methods (seen as successors of MDS).



\subsection{Extendable and invertible embedding with autoencoders}

The AE formulation presented (\ref{eq:AE}) departs  from manifold learning approaches as it lacks an explicit condition to recover geometrical interactions between observations. To fill that gap, we propose a general framework called GRAE (Geometry regularized autoencoders) which explicitly penalizes misguided representations in the latent space from a geometric perspective. Thus, we add a soft constraint in the bottleneck of the autoencoder as follows: 
\begin{align}
\label{eq:GRAE}
    \operatorname*{arg\ min}_{f,f^{\dagger}}\mathcal{L}(f,f^{\dagger}) = \mathcal{L}_{r}(X, f^{\dagger}(f(X))) + \lambda\mathcal{L}_{g}(f(X), \mathcal{E}).
\end{align}
%
%
%
%
%
%
The $\mathcal{L}_g$ term in (\ref{eq:GRAE}) is the geometric loss, penalizing the discrepancy between the latent representation and the embedding $\mathcal{E}$ previously learned by a manifold learning algorithm. Specifically, given an embedding of training points $\mathcal{E} = \{e_{1}, e_{2}, \ldots, e_{n}\}$, we define the geometric loss as $\mathcal{L}_{g}(f(X), \mathcal{E}) =  \sum_{i=1}^{n} \|e_{i} - f(x_{i})\|^{2}$.

The parameter $\lambda\geq 0$ determines how strongly the latent space of the AE should match the embedding $\mathcal{E}$. Thus the network will implicitly force the latent space of the autoencoder to preserve the relationships learned by the manifold learning technique, resulting in a non-linear embedding function $f$ and its inverse $f^\dagger$ that are consistent with sensible geometric properties.  This regularization can be applied with any manifold learning approach, whether it be Isomap, t-SNE, etc. The resulting latent space will then inherit the corresponding strengths and weaknesses of the selected approach.

\subsection{Geometric regularization with diffusion manifold learning}

To generate $\mathcal{E}$ in this work,  we use PHATE~\cite{moon2019visualizing} as it has proven to preserve long-range relationships (global structure) in a low-dimensional representation beyond the capabilities of spectral methods such  as Laplacian eigenmaps, Diffusion Maps, LLE, and Isomap, especially when the dimension $d$ is required to be 2 or 3 for visualization. PHATE is built upon diffusion geometry \cite{coifman2006diffusion,nadler2006diffusion}. PHATE first computes an $\alpha-$decay kernel with an adaptive bandwidth, which captures local geometry while remaining robust to density changes. The kernel matrix is normalized to obtain a probability transition matrix $P$ (diffusion operator) between every pair of points. Various scales of the geometry can then be uncovered by computing a $t$-step random walk over $P$, with a higher $t$ implying more diffusion, pushing transition probabilities to a more global scale.















The parameter $t$ is automatically chosen by studying the entropy of a distribution over the eigenvalues of $P$ (known as the von Neumann Entropy) as $t$ increases. Typically, the first few $t$-steps  lead to a sharp drop in the entropy, which is thought in \cite{moon2019visualizing}  to be the process of denoising the transition probabilities, whereas later steps will reduce entropy at a lower rate, thus, slowly losing meaningful information in the structure. Subsequently PHATE computes the so-called potential distances $D'_{t}$, which have proven to be adequate distances between the transition probabilities encoded in $P^{t}$. Finally, metric MDS is applied to $D'_{t}$ to optimally preserve the potential distances in a low-dimensional representation. Fig. ~\ref{fig:diagram} shows an overview of GRAE using the PHATE embedding.

\begin{figure}[!b]
    \centering
    \includegraphics[width=0.45\textwidth,trim=45pt 80pt 82pt 25pt,clip]{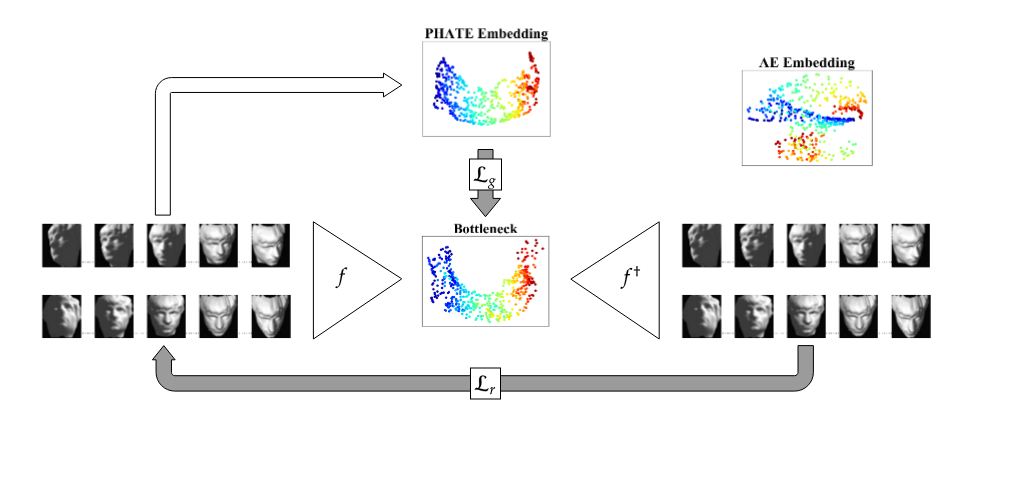}
    \caption{\textbf{Overview of GRAE on the Faces dataset~\cite{tenenbaum2000global}}. Geometric regularization is applied to enforce similarity between GRAE and PHATE embeddings. The vanilla AE embedding (top right) is included for reference.}
    \label{fig:diagram}
\end{figure}


\subsection{$\lambda$ parameter updating} 
While we expect $\mathcal{E}$ to be a good reference for the latent space on most problems, the ideal value for $\lambda$ can vary significantly depending on the dataset. Indeed, setting $\lambda$ too low  prevents the encoder from learning a good geometry while setting it too high might be detrimental to reconstruction quality.  To avoid the need to tune $\lambda$, we developed an update schedule for the parameter $\lambda$, which is set to a high value at the beginning of training and is progressively relaxed as follows:
\begin{align}
\label{eq:schedule}
    \lambda \leftarrow \frac{-\lambda_{\text{max}}\text{exp}((epoch-N_{e}/2)*\alpha)}{1+ \text{exp}((epoch-N_{e}/2)*\alpha)} + \lambda_{\text{max}},
\end{align}
where $\alpha$ can be adjusted to change the steepness of the relaxing curve. The bigger its value, the closer the update schedule resembles a step function. In practice we set $\alpha = 0.2$. $N_{e}$ and $\lambda_{\text{max}}$ are the total number of epochs and the initial $\lambda$ value, respectively. Roughly speaking, this schedule allows the AE to focus on good latent space geometry during the first half of training and then refine the embedding according to reconstruction imperatives during the second half. We observe empirically that this approach is more robust than using a constant $\lambda$ value, as illustrated in Fig.~\ref{fig:schedule}.

\begin{figure}[!t]
    \centering
    \includegraphics[width=0.4\textwidth]{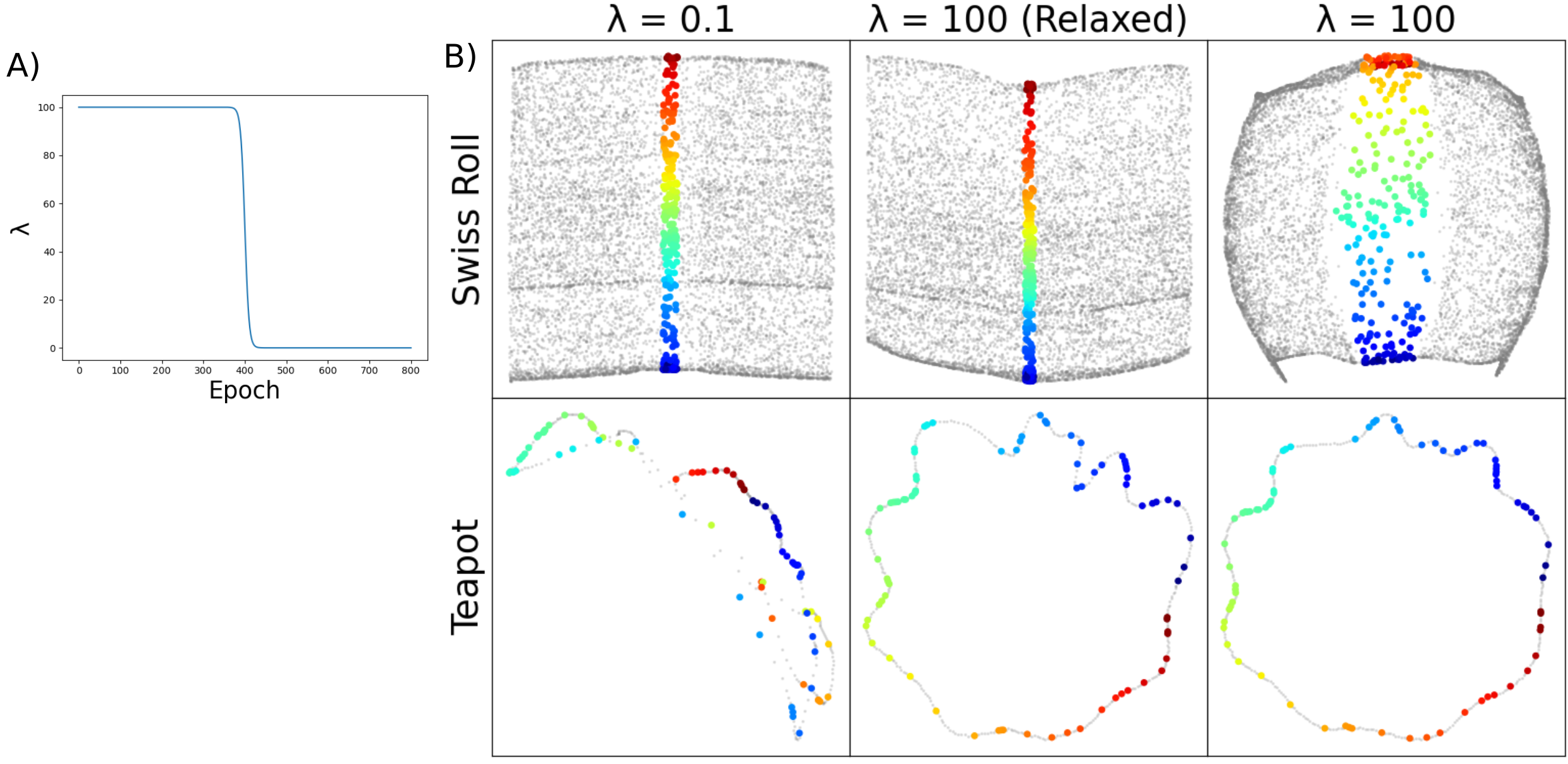}
    \caption{\textbf{$\lambda$ relaxation.} \textbf{A)} Plot of  (\ref{eq:schedule}).  \textbf{B)} GRAE latent space visualizations with three different $\lambda$ schemes on two problems~: the Swiss Roll and the Teapot (see subsection \ref{subsec:setup} for details), where we expect dimensionality reduction methods to respectively uncover a rectangular plane and a circular structure. We observe that the optimal $\lambda$ value is low for Swiss Roll (left) and high for Teapot (right). Relaxing $\lambda$ (center) allows us to recover the expected geometry on both problems with the same hyperparameter value. }
    \label{fig:schedule}
\end{figure}











\subsection{Embedding-based optimization}

Most methods introduced in Section \ref{sec:related_work} rely on jointly minimizing the reconstruction loss and the regularizing loss during training. We instead choose to precompute $\mathcal{E}$  for three reasons. First, typical optimization objectives based on manifold learning, such as those used by Diffusion Nets, EAER and parametric t-SNE, require a pairwise affinity matrix which needs to be accessible during training and entails a memory cost of scale $\mathcal{O}(n^2)$. While our approach needs a similar matrix to precompute $\mathcal{E}$, only the latter needs to be retained for training the AE architecture, with a memory requirement of $\mathcal{O}(nd)$ where $d$ is the latent space dimension and typically $d << n$. Other methods, such as TAE, do not require computing such a matrix on the whole dataset, but instead rely on computing distance matrices over mini-batches. Given a mini-batch B of size $p$, this approach leads to $\mathcal{O}(p^2)$ additional operations for every gradient step when compared to typical AE training. The GRAE loss, for its part, only requires computing the euclidean distance between $x_i \in \mathcal{B}$ and its target $e_i \in \mathcal{E}$, which is $\mathcal{O}(p)$. Overall, this optimization approach is better for big data applications given the complexity of competing methods.

Second, the  optimization techniques employed for some dimensionality reduction methods can be superior in performance than a neural network optimization. For instance, PHATE uses the SMACOF algorithm to perform MDS over the potential distances matrix. In practice, we find that this finds a better optimum than stochastic gradient descent or its variants. 

Finally, we depart from previous methods by providing a more general approach. In many applications, there may not be strong reasons for imposing a particular relationship in the geometric loss that resembles a loss function from a specific kernel method. Any approach employed to find $\mathcal{E}$, whether it be PHATE, Isomap, t-SNE, LLE, etc., is already performing an optimization to its particular loss function, imposing the preservation of its desired geometric relationships in the data. Thus, GRAE will implicitly enforce such a relationship.

\subsection{Big data applications}

Here we present an extension of GRAE combined with PHATE that is especially suitable for managing large data (see Fig.~\ref{fig:procrustes_extension}). We propose to partition the data in mini-batches $\mathcal{B} = \{B_{1}, B_{2}, \ldots, B_{N}\}$ each  containing common observations $\bm{X_{c}}$ and unique observations $\bm{X_{i}}$. Thus,  $B_{i} = \{\bm{X_{c}} \cup \bm{X_{i}}\}$. 
Then, we apply PHATE to each mini-batch, which produces an embedding $\mathcal{E}_{i}= \{\mathcal{E}^{\bm{X_{c}}} \cup \mathcal{E}^{\bm{X_{i}}}\}$.
Since each embedding might vary in orientation, scale and reflection with respect to the others, we apply the Procrustes method \cite{wang2008manifold} among the common points to extract a linear transformation, which is then applied to the whole mini-batch embedding. This allows us to consistently combine all embeddings. Clearly, such final embedding is not as refined as computing PHATE for the whole dataset, and some local information is lost. Fortunately, GRAE  is able to refine the discrepancies, generating near similar embeddings whether $\mathcal{E}$ is generated by computing PHATE over the whole data, or by the Procrustes transformations over mini-batches. This makes the computation complexity linear with respect to the number of mini-batches.    

\begin{figure}[!t]
    \centering
    \includegraphics[width=0.49\textwidth]{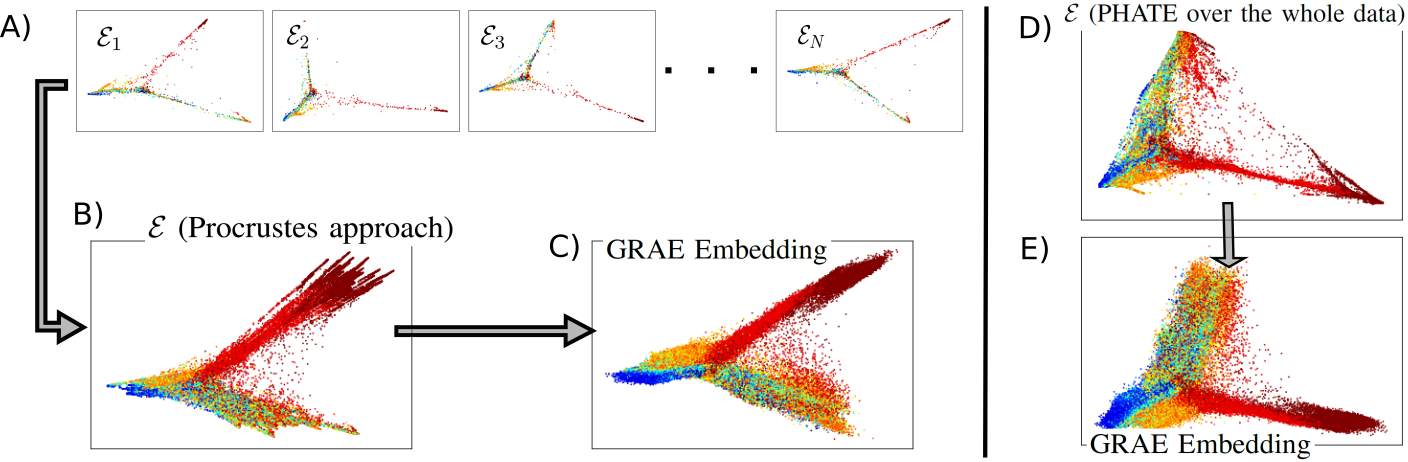}
    \caption{\textbf{Scalable GRAE.} Overview of GRAE applied to 200,000 observations of iPSC data (see Section \ref{subsec:datasets}). \textbf{A)} Multiple mini-batch embeddings, each of which share some common observations. \textbf{B)} Combined embedding using the Procrustes method to align the mini-batch embeddings. \textbf{C)} GRAE embedding using (\textbf{B}) as $\mathcal{E}$ in the geometric loss. \textbf{D)} PHATE embedding computed over the whole data set.  \textbf{E)} GRAE embedding using (\textbf{D}) as $\mathcal{E}$ in the geometric regularization. Although both approaches produce near similar embeddings, the mini-batch approach takes around 850 seconds to be computed and scales linearly. In contrast,  PHATE applied to the whole data takes around 3894 seconds and scales quadratically.}
    \label{fig:procrustes_extension}
\end{figure}

\section{Results}
\label{sec:results}

In this section, we experimentally compare GRAE\footnote{Source code is available at \url{https://github.com/KevinMoonLab/GRAE}} with a standard AE, Diffusion nets~\cite{mishne2019diffusion}, TAE~\cite{moor2019topological}, EAER-Margin~\cite{yu2013embedding} and UMAP~\cite{UMAP2018}  on 6 different problems using a two-dimensional latent space. The motivation behind such a low-dimensional latent space is two-fold : to visualize the geometry of the latent space, and to offer a challenging dimensionality reduction task. Training is  unsupervised  and ground truths are only used for visualizing and scoring embeddings.

\subsection{Experimental setup}
\label{subsec:setup}
\label{subsec:datasets}

\paragraph{Compared methods}

A comparison with a standard AE was required to measure improvements in latent space geometry and reconstruction quality. We selected the remaining regularized autoencoders (EAER-Margin\footnote{While EAER also introduces loss terms based on Laplacian Eigenmaps and MDS, the margin-based loss performed best according to their benchmarks.}, TAE and Diffusion Nets) based on the similar goal of their objective functions and the geometric loss defined in this work, which is to introduce more structure in the latent space via a prior driven by geometry or topology. While other  contributions have been made to improve representation learning with autoencoders, we do not think they are directly comparable to our approach nor relevant given our experimental setup. Some methods, such as Denoising and Sparse Autoencoders, do not explicitly regularize the structure of the latent space, which limits their utility for data exploration and visualization. In fact, they can be used in tandem with our geometric loss (and other compared methods) since their  objectives are unlikely to conflict with the goal of achieving good latent space geometry. Other methods that do impose a prior on the latent space can be unsuitable for general manifold learning. For example, the VAE Gaussian prior, while useful for generative sampling, is inadequate to model the true manifold of many real-life datasets.

We finally include UMAP~\cite{UMAP2018} as a baseline to compare our approach to a ``pure'' manifold learning algorithm. This choice is motivated by the presence of a native extension to new points in UMAP's implementation as well as an inverse transform from the latent space back to the input space based on linear combinations of nearest neighbors, thus offering a natural comparison to autoencoders over all the relevant metrics. We did not benchmark parametric t-SNE \cite{van2009learning} given the absence of an invertible mapping and the fact that  UMAP is similar to t-SNE from an algorithmic standpoint and provides qualitatively similar embeddings at a lower computational cost.

\paragraph{Datasets}

We perform comparisons on 6 datasets. The first one is the classic manifold problem known as ``Swiss Roll'' where data points are lying on a two dimensional plane ``rolled'' in a three dimensional ambient space. Classical approaches such as PCA or MDS typically fail to recover the non linear geometry of the data, as they rely on pairwise Euclidean distances instead of the true geodesic distances along the curvature of the roll. We generated 10,000 points on the Swiss Roll using the \texttt{scikit-learn} library. The manifold is then stretched and rotated to achieve unit variance.

Two datasets focus on full object rotations, where samples lie on circular manifolds. One problem is derived from the MNIST dataset~\cite{lecun2010mnist}, where three digits are picked randomly and rotated 360 times at one-degree intervals, for a total of 1080 images. Another one is known as the Teapot problem \cite{weinberger2004learning}, in which 400 RGB images of size 76 x 128 feature a rotating textured teapot.

We benchmark methods on two additional image datasets. The first one ("Object Tracking") was created with a 16~x~16 small character moving on a 64~x~64 background. Approximately 2000 RGB images were generated and Gaussian noise was added to the background to increase difficulty. The intrinsic manifold consists of the plane spanned by the x and y character coordinates on the background. The second is known as the UMIST Faces dataset \cite{graham1998characterising}, where different views (over a 90$^{\circ}$ interval) of the faces of 20 different subjects are shown in 575 gray scale images of size 112 x 92.

The final dataset consists of single-cell mass cytometry data showing iPSC reprogramming of mouse
embryonic fibroblasts (hereinafter, "iPSC") as introduced in \cite{zunder2015continuous}. The data show the expression of 33 markers in 220,450 embryonic cells at various stages of development. We know from \cite{moon2019visualizing} that cells, while initially similar, eventually specialize into two different groups, leading to a two-branch Y-shaped manifold. The only known ground truth in this case is the age of the cell when measured.

\paragraph{Architecture and hyperparameters}

We use the same AE architecture for all autoencoder-based models, which consists, in the case of Swiss Roll and iPSC, of 3 fully-connected hidden layers in the encoder and in the decoder with a 2D latent space, producing the following sequence of neurons: 800-400-200-2-200-400-800. For the image datasets, the first layer is replaced by two convolutional layers with max pooling and the last layer of the decoder is substituted for two deconvolution layers.  We apply ReLU activations on all of the layers except in the bottleneck and the output of the decoder. To prevent over-fitting, we include an L2 penalization with a \textit{weight-decay} parameter equal to 1. We use a learning rate of .0001 and a batch size of 128, except for Diffusion Nets, which do not support mini-batches. Neural networks are trained for 200 epochs on the Swiss Roll and iPSC data and for 800 epochs on remaining datasets. MSE is used for the reconstruction error in the objective function. For GRAE, $\lambda_{max}$ was set to 100. Similarly, $\lambda$ was set to 100 for TAE and EAER-Margin. We also used a 100 multiplier on the eigenvector constraint of Diffusion Nets, as suggested in their paper. As for hyperparameters specific to PHATE (for computing the GRAE $\mathcal{E}$), UMAP and Diffusion maps (for Diffusion Nets), they were minimally tuned for each problem to achieve decent 2D visualizations.\footnote{We note however, that tuning Diffusion maps on some problems, most notably UMIST, proved to be much harder than tuning PHATE or UMAP.}


\subsection{Qualitative evaluation}
\label{subsec:qualitative}

\begin{figure}[!ht]
    \centering
    \includegraphics[width=.485\textwidth]{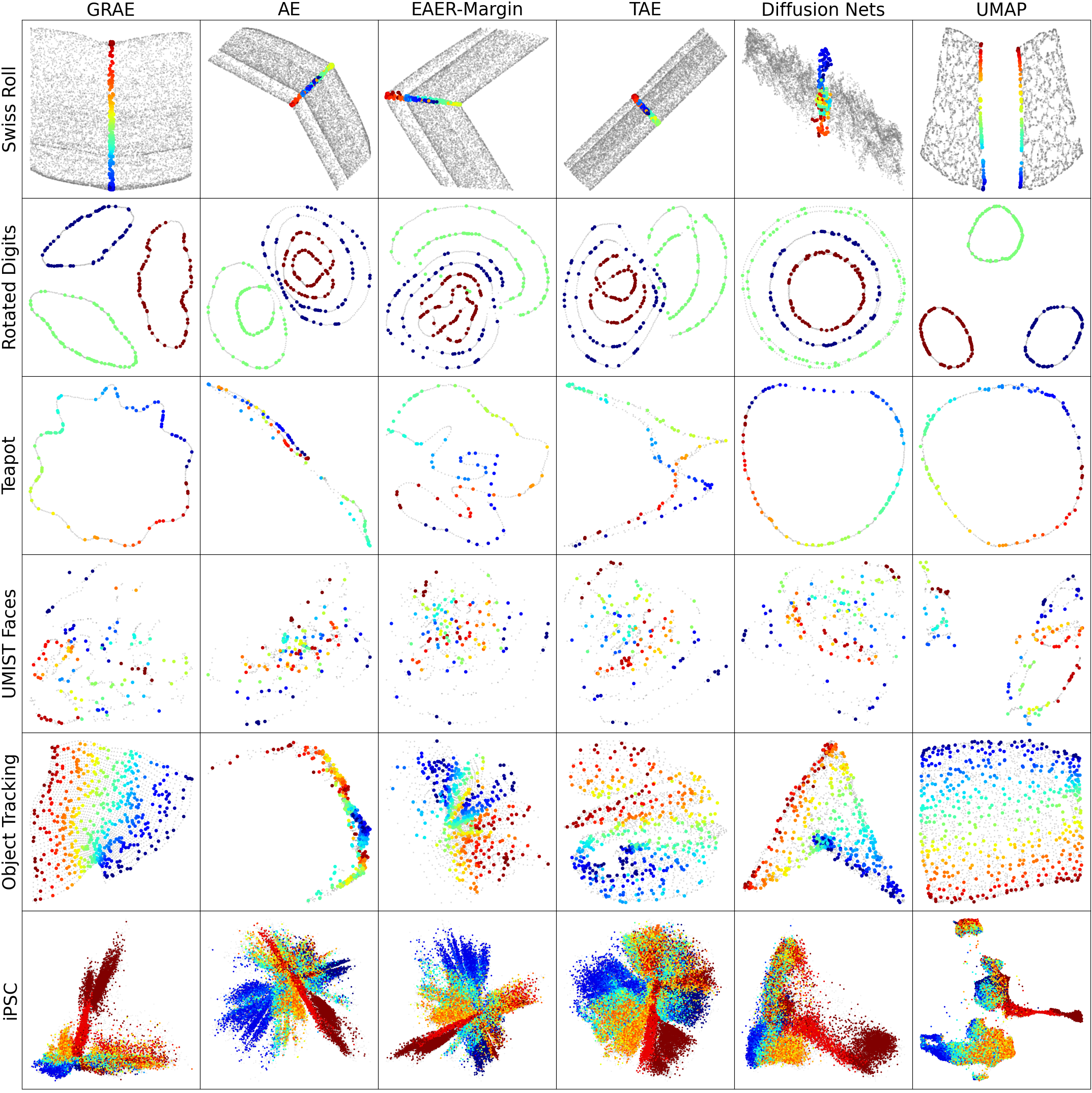}
    \caption{\textbf{Latent space visualizations.} Latent representations learned in a fully unsupervised manner by all considered methods on 6 datasets. Training points are grey scale and testing points are colored according to a known ground truth factor. Only GRAE recovers a sensible geometry for all problems.}
    \label{fig:plots}
\end{figure}

We first qualitatively evaluate GRAE and the other methods by visualizing the embedding layer after training as seen in in Fig.~\ref{fig:plots}. We first notice that GRAE recovers a sensible geometry for all problems while other methods fail at basic tasks such as uncoiling the Swiss Roll, disentangling the Rotated Digits, showing the coordinate plane on Object Tracking or recovering a circular structure on Teapot.
Only GRAE and Diffusion Nets show the two expected branches of the iPSC manifold. UMAP also shows good geometry in general, but tears the overall structure of the manifold on the Swiss Roll and iPSC problems despite their known continuous natures. TAE and EAER-Margin generally show embeddings similar to those produced by the AE, albeit with more sparsity. Finally, no method shows a clear global structure on the UMIST Faces problem, which is expected given that images were taken from 20 different subjects, but we still note that GRAE and UMAP demonstrate better local structure.

\subsection{Quantitative evaluation}
\label{subsec:quantitative}


\begin{table}[htbp]
\caption{Quantitative Evaluations of Methods}

\centering
\resizebox{.49\textwidth}{!}{
\begin{tabular}{|c|c|c|c|r|}
\hline &     &  \multicolumn{3}{|c|}{\textbf{Metrics}}\\
\cline{3-5} 
\textbf{Dataset} & \textbf{Model} & \textbf{\textit{$\boldsymbol{R^2}$}} & \textbf{\textit{ MSE}} & \textbf{\textit{Rel. MSE}} \\
\hline


\multirow{6}{*}{Swiss Roll} & GRAE &  \color[HTML]{2F781E}{\ul \textbf{0.9762 (1)}} &  \color[HTML]{2F781E}{\ul \textbf{0.0034 (1)}} &  \color[HTML]{2F781E}{\ul \textbf{-83.8 \% (1)}} \\
     & AE &                 \color[HTML]{A5A600}0.5136 (3) &                 \color[HTML]{C69B00}0.0210 (4) &                   \color[HTML]{C69B00}0.0 \% (4) \\
     & EAER-Margin &                 \color[HTML]{D70b0B}0.3746 (6) &                 \color[HTML]{D87D00}0.0246 (5) &                  \color[HTML]{D87D00}17.1 \% (5) \\
     & TAE &                 \color[HTML]{C69B00}0.4905 (4) &                 \color[HTML]{A5A600}0.0196 (3) &                  \color[HTML]{A5A600}-6.7 \% (3) \\
     & Diffusion Nets &                 \color[HTML]{D87D00}0.4257 (5) &                 \color[HTML]{D70b0B}0.0546 (6) &                 \color[HTML]{D70b0B}160.0 \% (6) \\
     & UMAP &        \color[HTML]{699010}\textbf{0.8455 (2)} &        \color[HTML]{699010}\textbf{0.0042 (2)} &        \color[HTML]{699010}\textbf{-80.0 \% (2)} \\
\hline
\multirow{6}{*}{Rotated Digits} & GRAE &        \color[HTML]{699010}\textbf{0.9829 (2)} &  \color[HTML]{2F781E}{\ul \textbf{0.0027 (1)}} &  \color[HTML]{2F781E}{\ul \textbf{-69.3 \% (1)}} \\
     & AE &                 \color[HTML]{A5A600}0.3757 (3) &                 \color[HTML]{A5A600}0.0088 (3) &                   \color[HTML]{A5A600}0.0 \% (3) \\
     & EAER-Margin &                 \color[HTML]{D87D00}0.3181 (5) &        \color[HTML]{699010}\textbf{0.0061 (2)} &        \color[HTML]{699010}\textbf{-30.7 \% (2)} \\
     & TAE &                 \color[HTML]{C69B00}0.3359 (4) &                 \color[HTML]{C69B00}0.0101 (4) &                  \color[HTML]{C69B00}14.8 \% (4) \\
     & Diffusion Nets &                 \color[HTML]{D70b0B}0.2530 (6) &                 \color[HTML]{D87D00}0.0300 (5) &                 \color[HTML]{D87D00}240.9 \% (5) \\
     & UMAP &  \color[HTML]{2F781E}{\ul \textbf{0.9845 (1)}} &                 \color[HTML]{D70b0B}0.0653 (6) &                 \color[HTML]{D70b0B}642.0 \% (6) \\
\hline
\multirow{6}{*}{Teapot} & GRAE &  \color[HTML]{2F781E}{\ul \textbf{0.9989 (1)}} &        \color[HTML]{699010}\textbf{0.0032 (2)} &        \color[HTML]{699010}\textbf{-62.4 \% (2)} \\
     & AE &                 \color[HTML]{D70b0B}0.2079 (6) &                 \color[HTML]{C69B00}0.0085 (4) &                   \color[HTML]{C69B00}0.0 \% (4) \\
     & EAER-Margin &                 \color[HTML]{C69B00}0.2526 (4) &  \color[HTML]{2F781E}{\ul \textbf{0.0027 (1)}} &  \color[HTML]{2F781E}{\ul \textbf{-68.2 \% (1)}} \\
     & TAE &                 \color[HTML]{D87D00}0.2287 (5) &                 \color[HTML]{D87D00}0.0097 (5) &                  \color[HTML]{D87D00}14.1 \% (5) \\
     & Diffusion Nets &                 \color[HTML]{A5A600}0.9933 (3) &                 \color[HTML]{A5A600}0.0038 (3) &                 \color[HTML]{A5A600}-55.3 \% (3) \\
     & UMAP &        \color[HTML]{699010}\textbf{0.9981 (2)} &                 \color[HTML]{D70b0B}0.0160 (6) &                  \color[HTML]{D70b0B}88.2 \% (6) \\
\hline
\multirow{6}{*}{UMIST Faces} & GRAE &                 \color[HTML]{A5A600}0.9371 (3) &  \color[HTML]{2F781E}{\ul \textbf{0.0092 (1)}} &  \color[HTML]{2F781E}{\ul \textbf{-35.7 \% (1)}} \\
     & AE &                 \color[HTML]{D70b0B}0.9040 (6) &                 \color[HTML]{D87D00}0.0143 (5) &                   \color[HTML]{D87D00}0.0 \% (5) \\
     & EAER-Margin &                 \color[HTML]{D87D00}0.9298 (5) &        \color[HTML]{699010}\textbf{0.0108 (2)} &        \color[HTML]{699010}\textbf{-24.5 \% (2)} \\
     & TAE &  \color[HTML]{2F781E}{\ul \textbf{0.9426 (1)}} &                 \color[HTML]{A5A600}0.0118 (3) &                 \color[HTML]{A5A600}-17.5 \% (3) \\
     & Diffusion Nets &        \color[HTML]{699010}\textbf{0.9407 (2)} &                 \color[HTML]{C69B00}0.0128 (4) &                 \color[HTML]{C69B00}-10.5 \% (4) \\
     & UMAP &                 \color[HTML]{C69B00}0.9348 (4) &                 \color[HTML]{D70b0B}0.0292 (6) &                 \color[HTML]{D70b0B}104.2 \% (6) \\
\hline
\multirow{6}{*}{Object Tracking} & GRAE &        \color[HTML]{699010}\textbf{0.9298 (2)} &  \color[HTML]{2F781E}{\ul \textbf{0.0410 (1)}} &   \color[HTML]{2F781E}{\ul \textbf{-6.6 \% (1)}} \\
     & AE &                 \color[HTML]{D70b0B}0.3658 (6) &                 \color[HTML]{D87D00}0.0439 (5) &                   \color[HTML]{D87D00}0.0 \% (5) \\
     & EAER-Margin &                 \color[HTML]{D87D00}0.4124 (5) &        \color[HTML]{699010}\textbf{0.0429 (2)} &         \color[HTML]{699010}\textbf{-2.3 \% (2)} \\
     & TAE &                 \color[HTML]{C69B00}0.4369 (4) &                 \color[HTML]{A5A600}0.0434 (3) &                  \color[HTML]{A5A600}-1.1 \% (3) \\
     & Diffusion Nets &                 \color[HTML]{A5A600}0.7806 (3) &                 \color[HTML]{C69B00}0.0435 (4) &                  \color[HTML]{C69B00}-0.9 \% (4) \\
     & UMAP &  \color[HTML]{2F781E}{\ul \textbf{0.9890 (1)}} &                 \color[HTML]{D70b0B}0.0855 (6) &                  \color[HTML]{D70b0B}94.8 \% (6) \\
\hline
\multirow{6}{*}{iPSC} & GRAE &        \color[HTML]{699010}\textbf{0.2620 (2)} &        \color[HTML]{699010}\textbf{0.7440 (2)} &          \color[HTML]{699010}\textbf{1.0 \% (2)} \\
     & AE &                 \color[HTML]{D70b0B}0.0919 (6) &  \color[HTML]{2F781E}{\ul \textbf{0.7366 (1)}} &    \color[HTML]{2F781E}{\ul \textbf{0.0 \% (1)}} \\
     & EAER-Margin &                 \color[HTML]{C69B00}0.1305 (4) &                 \color[HTML]{C69B00}0.7721 (4) &                   \color[HTML]{C69B00}4.8 \% (4) \\
     & TAE &                 \color[HTML]{D87D00}0.1296 (5) &                 \color[HTML]{A5A600}0.7603 (3) &                   \color[HTML]{A5A600}3.2 \% (3) \\
     & Diffusion Nets &                 \color[HTML]{A5A600}0.2571 (3) &                 \color[HTML]{D70b0B}1.1060 (6) &                  \color[HTML]{D70b0B}50.1 \% (6) \\
     & UMAP &  \color[HTML]{2F781E}{\ul \textbf{0.3609 (1)}} &                 \color[HTML]{D87D00}0.7741 (5) &                   \color[HTML]{D87D00}5.1 \% (5) \\
\hline

\end{tabular}

}


\label{tab:metrics}
\end{table}

\paragraph{Metrics}
We report a quantitive assessment of all considered methods in Table \ref{tab:metrics}. We score models on two tasks: i) recovering ground truth factors in a meaningful way in the latent space and ii) reconstructing samples from the latent space back to the input space adequately.

Reconstruction quality is assessed using the MSE  between the input and the reconstruction. Disentanglement of the latent factors is assessed by fitting a linear regression to predict said factors, using the embedding coordinates as regressors. High-quality embeddings should indeed be indicative of the data generating process and represent ground truth factors adequately, subject to a simple transformation.\footnote{For circular manifolds (Teapot, Rotated Digits), we center the manifolds before switching to polar coordinates and use the resulting angles as predictors. Furthermore, we align angle 0 of the ground truth and the embedded points to mitigate a "spin" in the embedding, which would break the linear relationship.} We report the resulting $R^2$ of the linear model to measure the strength of the relationship between the embeddings and a given ground truth factor.\footnote{On datasets with cluster structure (UMIST, Rotated Digits) where points are unlikely to lie on a connected manifold, we partition the embedding according to ground truth class labels, fit linear regressions independently on each part and report the average $R^2$ over all partitions.} If more than one factor exists, an $R^2$ score is computed for each factor and the average is shown. 

All metrics were computed on a test split to benchmark how well the considered methods generalize to new data points. On Swiss Roll, the test split consists of a thin middle slice of 250 points that is removed from the training split to study how various methods would behave when required to generalize to out-of-distribution data. On the remaining datasets, 20 \% of samples are sampled and set aside for testing. We report the average of 10 runs using different train-test splits.

\paragraph{Discussion}

 We report quantitative results in Table \ref{tab:metrics}. Echoing its convincing visualizations, GRAE quantitatively recovers the latent factors by being first or runner-up on the $R^2$ metrics on all problems. Only UMAP, a pure manifold learning algorithm, proves to be a stable competitor in that regard while other autoencoders with manifold learning or topological components generally fail to disentangle the latent factors. Overall, our results suggest that GRAE provides a more faithful representation with superior ability to reveal latent structure and variables in data.
 
 While GRAE, UMAP and Diffusion Nets do recover the known ground truth factor on the iPSC data (i.e. age of the cells) better than the AE, we note that this factor  offers an imperfect overview of the data manifold and many other latent factors are likely at play. Furthermore, no considered method manages to improve the AE reconstruction performance. A possible explanation is that gene measurements on single-cell data can be noisy, leading to a relatively high MSE even if the global structure of the manifold is correctly reconstructed (in other words, the decoder does not model the noise). We explore the matter of noise and the GRAE iPSC embedding in more depth in Section \ref{subsec:iPSC}.

With respect to the goal of providing invertible embeddings, some methods typically fail to consistently match the reconstruction error of vanilla AE, which is expected given that they do not specifically account for this metric when inverting latent coordinates (UMAP), or that they are driven by other objectives that may dominate their training over reconstruction terms  (TAE and Diffusion Nets). GRAE, on the other hand, surprisingly improves reconstruction on five benchmarks; most notably on the Swiss Roll, Rotated Digits, UMIST and Teapot. We provide further study of this property in the following subsection.

\paragraph{Performance} Some compared methods showed limitations while running our benchmarks, especially on larger problems like iPSC. The algorithm presented in EAER had to be improved to support mini-batch training. Such an improvement was not as straightforward for Diffusion Nets because of the eigenvector constraint, which requires the full dataset and the associated affinity matrix to be held in GPU memory for training. In this case, subsampling the training set on the Teapot, Object Tracking, and UMIST problems was necessary to run experiments on the same hardware\footnote{Experiments were executed using \texttt{Python} 3.6.9 and \texttt{Torch} 1.5.0 for deep learning components. We used author implementations for UMAP (0.4.3) and PHATE (1.0.4). We used our own implementations of EAER-Margin and Diffusion Nets. As for TAE, we reused the original source code for the topological soft constraint and adapted it to our AE architecture. We ran everything with an \textit{AMD Ryzen 5 3600 6-Core CPU @ 4.20GHz}, 32 GB of available RAM, and an \textit{NVIDIA GeForce 2600 Super} GPU with 8 GB of memory.} used to fit other methods. In the case of iPSC, in addition to subsampling, we had to fit Diffusion Nets on a CPU to meet the higher memory requirements. While TAE supports mini-batches natively, we noticed a large increase in computation time, making it by far the slowest method on all problems. 


\subsection{Impact of geometric regularization on reconstruction quality}
\label{subsec:reconstruction}

\begin{figure}[tb!]
    \centering
    \includegraphics[width = 0.49\textwidth]{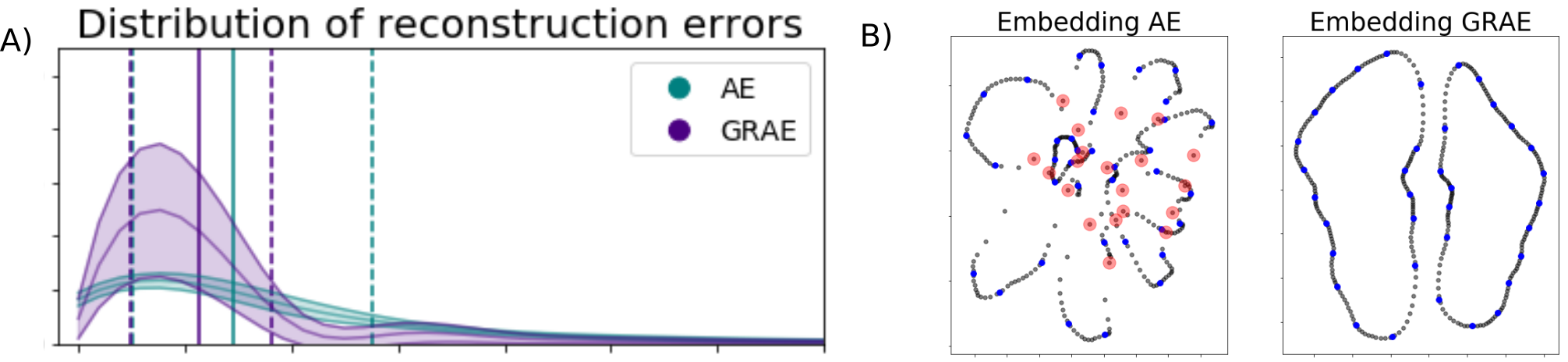}
    \caption{\textbf{Reconstructing latent space interpolations with GRAE.} \textbf{A)}~Distributions of errors of two rotated digits averaged over ten runs for  AE vs GRAE. Dashed lines represent the 1st and 3rd quartiles, and solid lines represent the median. We  notice that AE is more unstable than GRAE, having a heavier tail, since it fails completely to reconstruct certain images, while GRAE typically presents lighter tails. \textbf{B)} Typical embeddings produced for AE and GRAE. Blue points represent a subsample of the training data (subsampling only done for visualization purposes). Black points are the generated points on the latent space via interpolation. Red colored points in the AE embedding represent the 20 interpolated points with the highest reconstruction error. We observe that bad reconstruction typically occurs in sparse regions or crossing lines, i.e., in regions with poorly learned geometry.}
    \label{fig:errors_dist}
\end{figure}

\begin{figure}[t!]
    \centering
    \includegraphics[trim = 0 0 0 0, clip,width=0.5\textwidth]{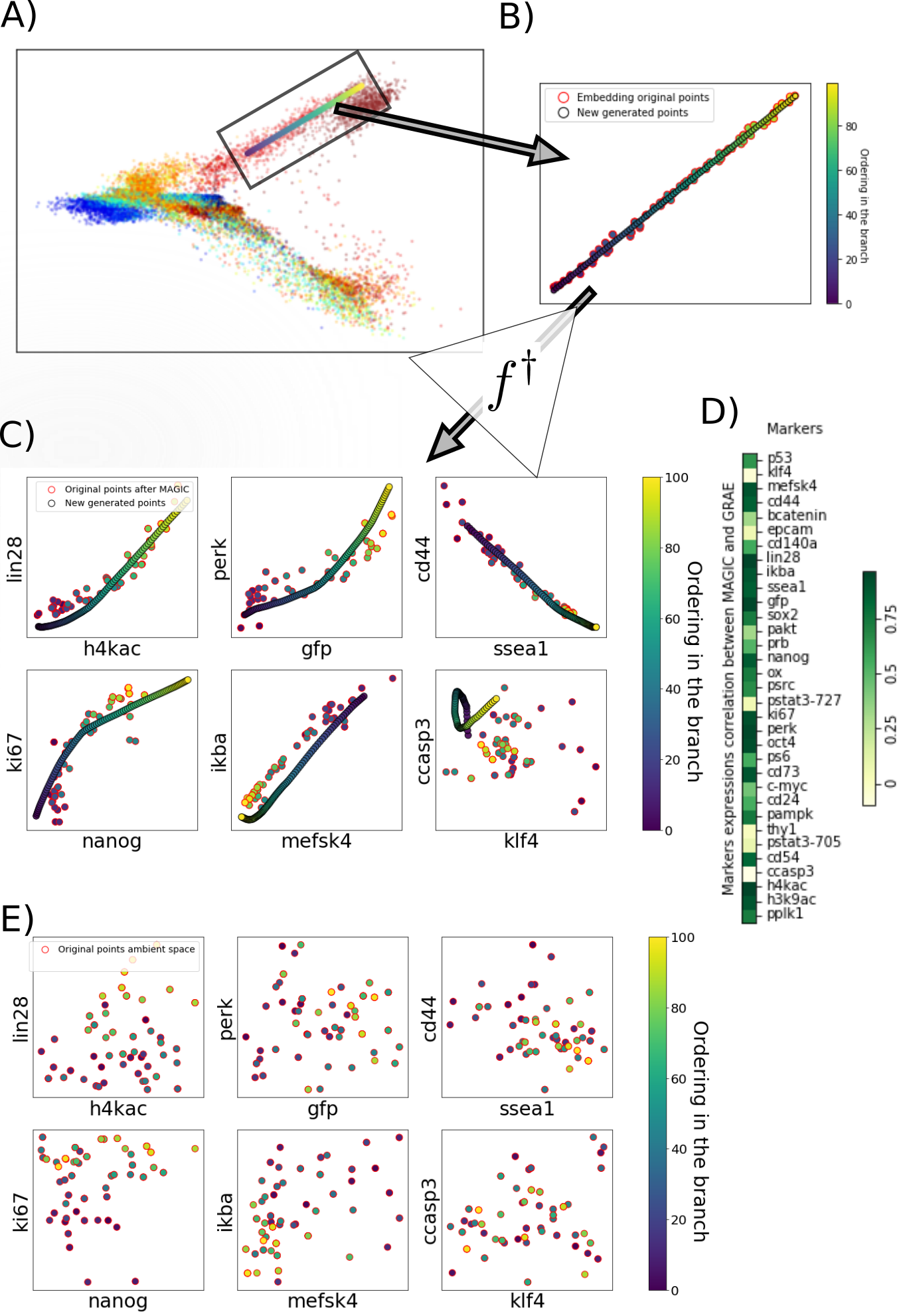}
    \caption{ \textbf{GRAE applied to the iPSC data} \cite{zunder2015continuous}. \textbf{A)} GRAE embedding of the iPSC data, and newly generated points across one of the branches. \textbf{B)} Newly generated points alongside their closest original observations in the embedding space. Both are colored by their ordering in the branch from the beginning of the branch to its tip. \textbf{C)} Pair plots for selected marker interactions in GRAE's reconstruction of the ambient space, compared against their closest original observations processed with MAGIC. Both show similar trends in most cases. \textbf{D)} Pearson correlation for each of the marker expressions between the data denoised by MAGIC and the GRAE reconstructions. Lack of correlation may indicate that neither GRAE nor MAGIC recovered a linear relationship. \textbf{E)} Pair plots in the input space of the raw original closest observations to the newly generated data. We  observe that there is no clear interaction between markers and no clear expression trend across the time-progression in the branch. }
    \label{fig:ipsc_exp}
\end{figure}


Based on the GRAE reconstruction errors (Table~\ref{tab:metrics}), we observe that GRAE improves the MSE of the decoder, despite adding a regularization term that deteriorates the reconstruction error global minimum. A possible explanation is that some latent space shifts caused by geometric regularization (e.g. forcing a circle on Teapot, uncoiling the Swiss Roll) actually drive gradient descent out of the local MSE minima in which vanilla AE falls. Indeed, most AE embeddings in Fig.~\ref{fig:plots} show structural overlaps where points from different regions of the original data manifold share the same latent space coordinates, meaning the encoder function $f$ is not injective and hence not invertible. By enforcing better geometry, GRAE appears to favor injective encoder mappings which, assuming bijectivity on the manifold, should facilitate learning of the inverse function $f^{\dagger}$, leading to the better reconstructions. 

Additionally, the regularization generates a more stable reconstruction over the whole dataset, especially in those regions where the autoencoder produces inaccurate geometry in the latent space. To support these claims, we conduct an experiment on two rotated MNIST digits (Fig.  \ref{fig:errors_dist}), generating a full rotation for each of the digits and removing in-between observations as the test set. After training GRAE on the remaining observations, we interpolate over consecutive observations in the embedding space i.e., consecutive angle rotations in the training set. Then we compute the reconstruction error between the generated points via interpolation with the previously removed in-between observations.
This experiment shows that learning accurate geometry from the data can be useful to generate new points on the latent space via interpolation over adjacent observations. Such points can then be fed to the decoder, generating a more faithfully reconstructed geodesic trajectory between points on the manifold in the ambient space in comparison to AE. 

\subsection{iPSC data case study}
\label{subsec:iPSC}

Finally, we present a case study showing the capabilities of GRAE to recover marker interactions in induced pluripotent stem cell (iPSC) mass cytometry data \cite{zunder2015continuous}. We show our experimental outline in Fig.~\ref{fig:ipsc_exp}. Mass cytometry data is  noisy (Fig.~\ref{fig:ipsc_exp}D) and marker expression interactions are difficult to extract from the raw data. This issue can be tackled with methods such as MAGIC \cite{van2018recovering}. MAGIC is a powerful diffusion-based approach that performs data imputation and denoising. It has been shown to be particularly useful to recover gene-gene interactions in complicated single-cell RNA-sequencing datasets. Thus, we choose to compare GRAE's reconstructed ambient space with the MAGIC transformation of the raw data. Figure~\ref{fig:ipsc_exp}(C,E) shows how GRAE is able to denoise the data, and recover gene markers interactions from the raw data in a similar way as MAGIC. We notice that large discrepancies between MAGIC and GRAE only occur when neither MAGIC nor GRAE  capture clear interactions, as shown in the bottom-right subplot in Fig.~\ref{fig:ipsc_exp}C.

\section{Conclusion}
\label{sec:conclusion}
We proposed geometry regularized autoencoder (GRAE), a general parametric framework to enhance autoencoders' latent representation by taking advantage of well-established manifold learning methods. By imposing a geometrical soft constraint on the bottleneck of the autoencoder, we demonstrated empirically how GRAE can achieve good visualizations and good latent representations  on several performance metrics compared to AE and other methods motivated by geometry. Furthermore, GRAE is equipped with an inverse mapping that often produces a better reconstruction than AE. While the primary focus of this work is on using PHATE embeddings to regularize the bottleneck, we leave to future work the study of other manifold learning algorithms as constraints for learning AE representations with better geometry and the benefits they bring in terms of visualizations, reconstruction, and data generation.



\balance
\bibliographystyle{IEEEtran}
\bibliography{ref}



%

\end{document}